\documentclass{article}

\usepackage{microtype}
\usepackage{graphicx}
\usepackage{subfigure}
\usepackage{capt-of}
\usepackage{booktabs} 

\usepackage{hyperref}

\usepackage{amssymb}
\usepackage{bm}
\usepackage{enumitem}
\usepackage{pbox}
\usepackage{adjustbox}
\usepackage[T1]{fontenc}
\usepackage{todonotes}
\usepackage{sidecap}
\sidecaptionvpos{figure}{c}

\usepackage[accepted]{icml2019}

\icmltitlerunning{Parameter-Efficient Transfer Learning for NLP}

\begin{document}

\twocolumn[
\icmltitle{Parameter-Efficient Transfer Learning for NLP}



\icmlsetsymbol{equal}{*}

\begin{icmlauthorlist}
\icmlauthor{Neil Houlsby}{goo}
\icmlauthor{Andrei Giurgiu}{goo,equal}
\icmlauthor{Stanis\l{}aw Jastrz\c{e}bski}{jag,equal}
\icmlauthor{Bruna Morrone}{goo}
\icmlauthor{Quentin de Laroussilhe}{goo}
\icmlauthor{Andrea Gesmundo}{goo}
\icmlauthor{Mona Attariyan}{goo}
\icmlauthor{Sylvain Gelly}{goo}
\end{icmlauthorlist}

\icmlaffiliation{goo}{Google Research}
\icmlaffiliation{jag}{Jagiellonian University}

\icmlcorrespondingauthor{Neil Houlsby}{neilhoulsby@google.com}

\icmlkeywords{NLP, Transfer Learning}

\vskip 0.3in
]



\printAffiliationsAndNotice{\icmlEqualContribution} 

\begin{abstract}
Fine-tuning large pre-trained models is an effective transfer mechanism in NLP. However, in the presence of many downstream tasks, fine-tuning is parameter inefficient: an entire new model is required for every task. As an alternative, we propose transfer with adapter modules. Adapter modules yield a compact and extensible model; they add only a few trainable parameters per task, and new tasks can be added without revisiting previous ones. The parameters of the original network remain fixed, yielding a high degree of parameter sharing. To demonstrate adapter's effectiveness, we transfer the recently proposed BERT Transformer model to $26$ diverse text classification tasks, including the GLUE benchmark. Adapters attain near state-of-the-art performance, whilst adding only a few parameters per task. On GLUE, we attain within $0.4\%$ of the performance of full fine-tuning, adding only $3.6\%$ parameters per task. By contrast, fine-tuning trains $100\%$ of the parameters per task.\footnote{Code at \url{https://github.com/google-research/adapter-bert}}
\end{abstract}

\section{Introduction}

Transfer from pre-trained models yields strong performance on many NLP tasks~\citep{dai2015,howard2018universal,radford2018improving}.
BERT, a Transformer network trained on large text corpora with an
unsupervised loss, attained state-of-the-art performance on text classification
and extractive question answering~\citep{devlin2018bert}.

In this paper we address the online setting, where tasks arrive in a stream.
The goal is to build a system that performs well on all of them, but without training an entire new model for every new task.
A high degree of sharing between tasks is particularly useful for applications such as cloud services,
where models need to be trained to solve many tasks that arrive from customers in sequence.
For this, we propose a transfer learning strategy that yields \emph{compact} and \emph{extensible} downstream models.
Compact models are those that solve many tasks using a small number of additional parameters per task.
Extensible models can be trained incrementally to solve new tasks, without forgetting previous ones.
Our method yields a such models without sacrificing performance.

\begin{figure}[t]
\centering
\includegraphics[width=0.95\linewidth]{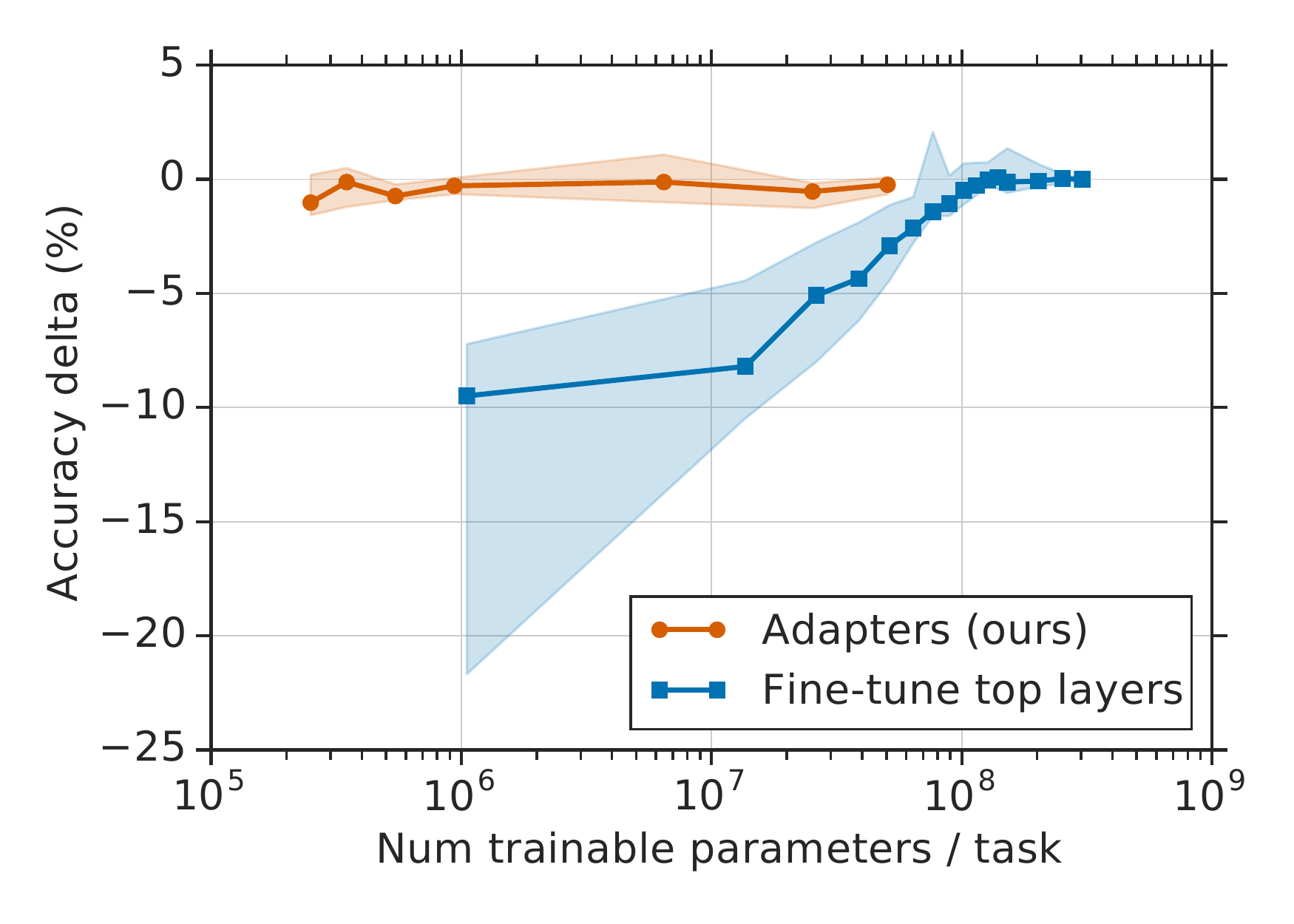}
\caption{
Trade-off between accuracy and number of trained task-specific parameters, for adapter tuning and fine-tuning.
The \emph{y}-axis is normalized by the performance of full fine-tuning, details in Section~\ref{sec:experiments}.
The curves show the $20$th, $50$th, and $80$th performance percentiles across nine tasks from the GLUE benchmark.
Adapter-based tuning attains a similar performance to full fine-tuning with two orders of magnitude fewer trained parameters.}
\label{fig:glue_summary_results}
\end{figure}

The two most common transfer learning techniques in NLP are feature-based transfer and fine-tuning.
Instead, we present an alternative transfer method based on adapter modules~\citep{rebuffi2017}.
Features-based transfer involves pre-training real-valued embeddings vectors.
These embeddings may be at the word~\citep{mikolov2013}, sentence~\citep{cer2019}, or paragraph level~\citep{le2014}.
The embeddings are then fed to custom downstream models.
Fine-tuning involves copying the weights from a pre-trained network and tuning them on the downstream task.
Recent work shows that fine-tuning often enjoys better performance than feature-based transfer~\citep{howard2018universal}.

Both feature-based transfer and fine-tuning require a new set of weights for each task.
Fine-tuning is more parameter efficient if the lower layers of a network are shared between tasks.
However, our proposed adapter tuning method is even more parameter efficient.
Figure~\ref{fig:glue_summary_results} demonstrates this trade-off.
The \emph{x}-axis shows the number of parameters trained per task;
this corresponds to the marginal increase in the model size required to solve each additional task.
Adapter-based tuning requires training two orders of magnitude fewer parameters to fine-tuning, while attaining similar performance.

Adapters are new modules added between layers of a pre-trained network.
Adapter-based tuning differs from feature-based transfer and fine-tuning in the following way.
Consider a function (neural network) with parameters $\bm w$: $\phi_{\bm w}(\bm x)$.
Feature-based transfer composes $\phi_{\bm w}$ with a new function, $\chi_{\bm v}$, to yield $\chi_{\bm v}(\phi_{\bm w}(\bm x))$.
Only the new, task-specific, parameters, $\bm v$, are then trained.
Fine-tuning involves adjusting the original parameters, $\bm w$, for each new task, limiting compactness.
For adapter tuning, a new function, $\psi_{\bm w, \bm v}(\bm x)$, is defined, where parameters $\bm w$ are copied over from pre-training.
The initial parameters $\bm v_0$ are set such that the new function resembles the original: $\psi_{\bm w, \bm v_0}(\bm x) \approx \phi_{\bm w}(\bm x)$.
During training, only $\bm v$ are tuned.
For deep networks, defining $\psi_{\bm w, \bm v}$ typically involves adding new layers to the original network, $\phi_{\bm w}$.
If one chooses $|\bm v|\ll|\bm w|$, the resulting model requires $\sim|\bm w|$ parameters for many tasks.
Since $\bm w$ is fixed, the model can be extended to new tasks without affecting previous ones.

Adapter-based tuning relates to \emph{multi-task} and \emph{continual} learning.
Multi-task learning also results in compact models.
However, multi-task learning requires simultaneous access to all tasks, which adapter-based tuning does not require.
Continual learning systems aim to learn from an endless stream of tasks.
This paradigm is challenging because networks forget previous tasks after re-training~\citep{mccloskey1989catastrophic,french1999catastrophic}.
Adapters differ in that the tasks do not interact and the shared parameters are frozen.
This means that the model has perfect memory of previous tasks using a small number of task-specific parameters.

We demonstrate on a large and diverse set of text classification tasks that adapters yield parameter-efficient tuning for NLP.
The key innovation is to design an effective adapter module and its integration with the base model.
We propose a simple yet effective, bottleneck architecture.
On the GLUE benchmark, our strategy almost matches the performance of the fully fine-tuned BERT,
but uses only 3\% task-specific parameters, while fine-tuning uses 100\% task-specific parameters.
We observe similar results on a further $17$ public text datasets, and SQuAD extractive question answering.
In summary, adapter-based tuning yields a single, extensible, model that attains near state-of-the-art performance in text classification.

\section{Adapter tuning for NLP}

\begin{SCfigure*}
\begin{tabular}{cc}
 \includegraphics[width=0.45\linewidth]{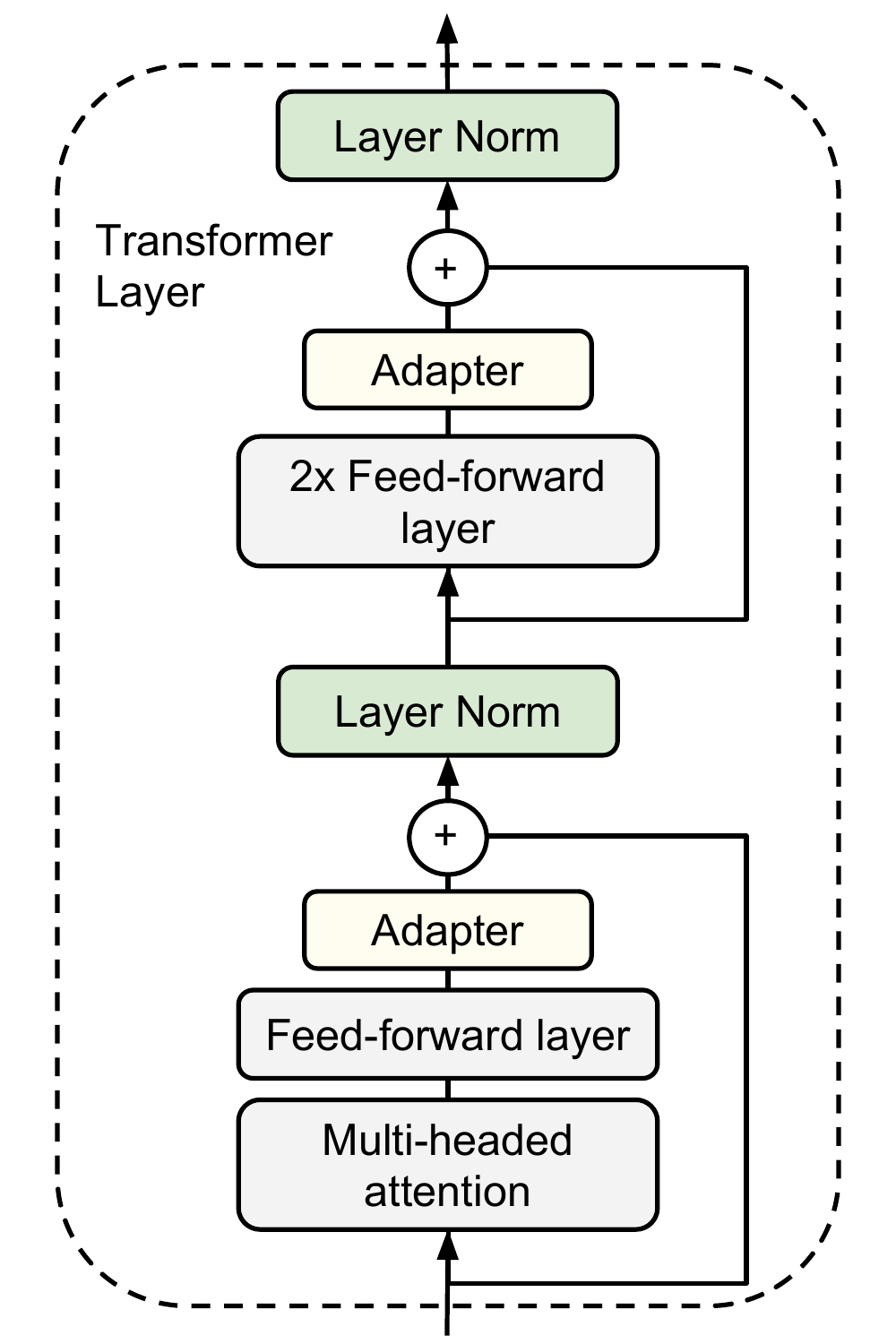}&
 \includegraphics[width=0.45\linewidth]{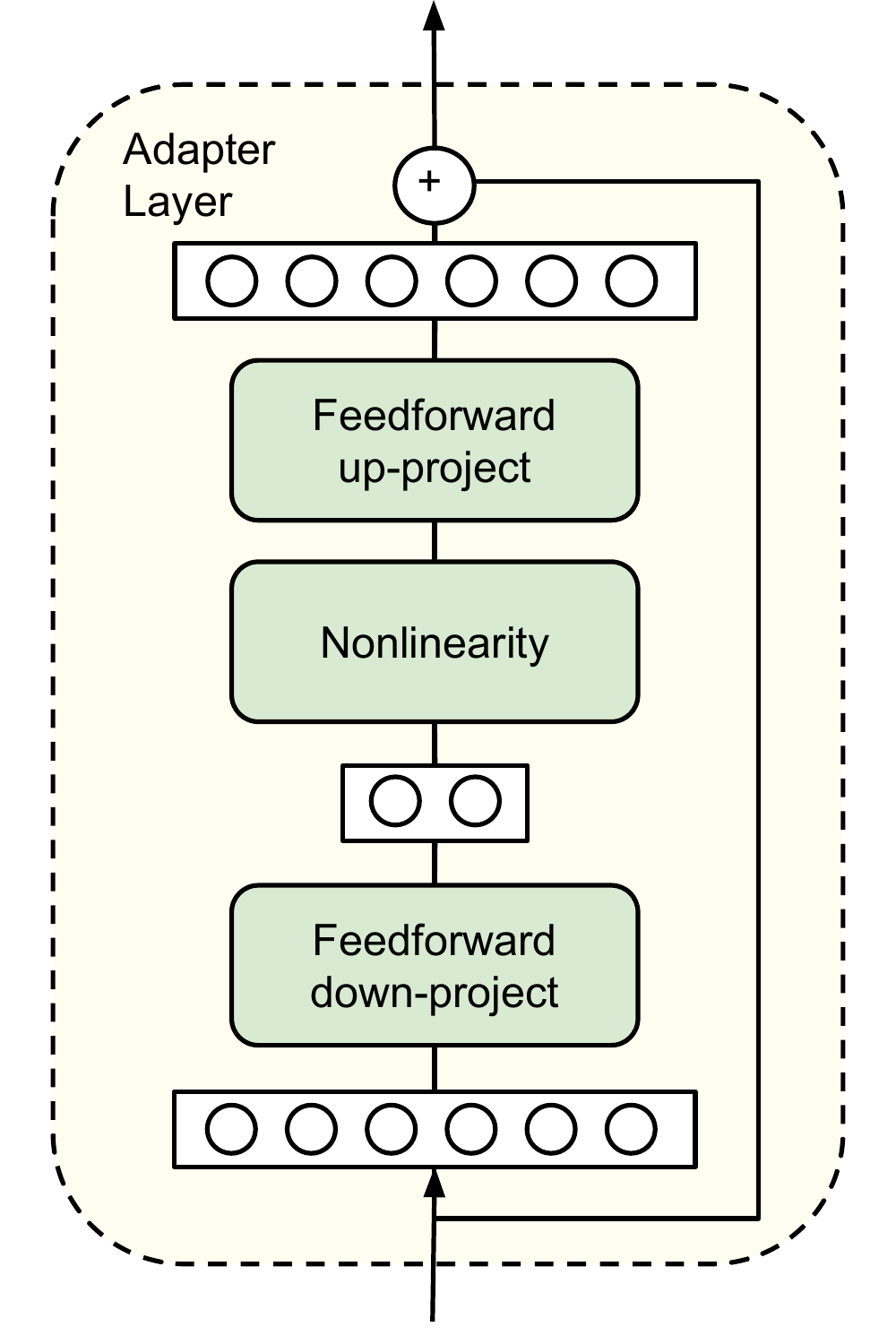}
\end{tabular}
 \caption{
 Architecture of the adapter module and its integration with the Transformer.
 \textbf{Left:} We add the adapter module twice to each Transformer layer:
 after the projection following multi-headed attention and after the two feed-forward layers.
 \textbf{Right:} The adapter consists of a bottleneck which contains few parameters relative to the attention and feedforward layers in the original model.
 The adapter also contains a skip-connection.
 During adapter tuning, the green layers are trained on the downstream data, this includes the adapter, the layer normalization parameters,
 and the final classification layer (not shown in the figure).
 \label{fig:adapters_transformer}}
\end{SCfigure*}

We present a strategy for tuning a large text model on several downstream tasks.
Our strategy has three key properties:
(i) it attains good performance,
(ii) it permits training on tasks sequentially, that is, it does not require simultaneous access to all datasets,
and (iii) it adds only a small number of additional parameters per task.
These properties are especially useful in the context of cloud services,
where many models need to be trained on a series of downstream tasks, so a high degree of sharing is desirable.

To achieve these properties, we propose a new bottleneck adapter module.
Tuning with adapter modules involves adding a small number of new parameters to a model, which are trained on the downstream task~\citep{rebuffi2017}.
When performing vanilla fine-tuning of deep networks, a modification is made to the top layer of the network.
This is required because the label spaces and losses for the upstream and downstream tasks differ.
Adapter modules perform more general architectural modifications to re-purpose a pre-trained network for a downstream task.
In particular, the adapter tuning strategy involves injecting new layers into the original network.
The weights of the original network are untouched, whilst the new adapter layers are initialized at random.
In standard fine-tuning, the new top-layer and the original weights are co-trained.
In contrast, in adapter-tuning, the parameters of the original network are frozen and therefore may be shared by many tasks.

Adapter modules have two main features: a small number of parameters, and a near-identity initialization.
The adapter modules need to be small compared to the layers of the original network.
This means that the total model size grows relatively slowly when more tasks are added.
A near-identity initialization is required for stable training of the adapted model;
we investigate this empirically in Section~\ref{sec:discussion}.
By initializing the adapters to a near-identity function, original network is unaffected when training starts.
During training, the adapters may then be activated to change the distribution of activations throughout the network.
The adapter modules may also be ignored if not required;
in Section~\ref{sec:discussion} we observe that some adapters have more influence on the network than others.
We also observe that if the initialization deviates too far from the identity function, the model may fail to train.

\subsection{Instantiation for Transformer Networks\label{sec:bottleneckadapter}}

We instantiate adapter-based tuning for text Transformers.
These models attain state-of-the-art performance in many NLP tasks,
including translation, extractive QA, and text classification problems~\citep{vaswani2017,radford2018improving,devlin2018bert}.
We consider the standard Transformer architecture, as proposed in~\citet{vaswani2017}.

Adapter modules present many architectural choices.
We provide a simple design that attains good performance.
We experimented with a number of more complex designs, see Section~\ref{sec:discussion},
but we found the following strategy performed as well as any other that we tested, across many datasets.

Figure~\ref{fig:adapters_transformer} shows our adapter architecture, and its application it to the Transformer.
Each layer of the Transformer contains two primary sub-layers: an attention layer and a feedforward layer.
Both layers are followed immediately by a projection that maps the features size back to the size of layer's input.
A skip-connection is applied across each of the sub-layers.
The output of each sub-layer is fed into layer normalization.
We insert two serial adapters after each of these sub-layers.
The adapter is always applied directly to the output of the sub-layer, after the projection back to the input size,
but before adding the skip connection back.
The output of the adapter is then passed directly into the following layer normalization.

To limit the number of parameters, we propose a bottleneck architecture.
The adapters first project the original $d$-dimensional features into a smaller dimension, $m$, apply a nonlinearity, then project back to $d$ dimensions.
The total number of parameters added per layer, including biases, is $2md+d+m$.
By setting $m\ll d$, we limit the number of parameters added per task;
in practice, we use around $0.5-8\%$ of the parameters of the original model.
The bottleneck dimension, $m$, provides a simple means to trade-off performance with parameter efficiency.
The adapter module itself has a skip-connection internally.
With the skip-connection, if the parameters of the projection layers are initialized to near-zero,
the module is initialized to an approximate identity function.

Alongside the layers in the adapter module, we also train new layer normalization parameters per task.
This technique, similar to conditional batch normalization~\citep{de2017modulating},
FiLM~\citep{perez2018}, and self-modulation~\citep{chen2019}, also yields parameter-efficient adaptation of a network; with only $2d$ parameters per layer.
However, training the layer normalization parameters alone is insufficient for good performance,
see Section~\ref{sec:param_efficiency}.

\section{Experiments\label{sec:experiments}}

We show that adapters achieve parameter efficient transfer for text tasks.
On the GLUE benchmark~\citep{wang2018glue},
adapter tuning is within $0.4\%$ of full fine-tuning of BERT, but it adds only $3\%$ of the number of parameters trained by fine-tuning.
We confirm this result on a further $17$ public classification tasks and SQuAD question answering.
Analysis shows that adapter-based tuning automatically focuses on the higher layers of the network.

\subsection{Experimental Settings}

We use the public, pre-trained BERT Transformer network as our base model.
To perform classification with BERT, we follow the approach in~\citet{devlin2018bert}.
The first token in each sequence is a special ``classification token''.
We attach a linear layer to the embedding of this token to predict the class label.

Our training procedure also follows~\citet{devlin2018bert}.
We optimize using Adam~\citep{kingma2014adam},
whose learning rate is increased linearly over the first $10\%$ of the steps, and then decayed linearly to zero.
All runs are trained on $4$ Google Cloud TPUs with a batch size of $32$.
For each dataset and algorithm, we run a hyperparameter sweep and select the best model according to accuracy on the validation set.
For the GLUE tasks, we report the test metrics provided by the submission website\footnote{\url{https://gluebenchmark.com/}}.
For the other classification tasks we report test-set accuracy.

We compare to fine-tuning, the current standard for transfer of large pre-trained models,
and the strategy successfully used by BERT.
For $N$ tasks, full fine-tuning requires $N{\times}$ the number of parameters of the pre-trained model.
Our goal is to attain performance equal to fine-tuning, but with fewer total parameters, ideally near to $1{\times}$.

\subsection{GLUE benchmark\label{sec:glue}}

\begin{table*}[t]
\centering
\begin{adjustbox}{max width=\textwidth}
\begin{tabular}{l|ll|rrrrrrrrr|r}
\toprule
{} & \pbox{3cm}{Total num\\ params} & \pbox{3cm}{Trained \\ params / task} & CoLA &     SST &    MRPC &   STS-B &     QQP & MNLI\textsubscript{m} & MNLI\textsubscript{mm} &    QNLI &     RTE &   Total \\
\midrule
BERT\textsubscript{LARGE} &  $9.0\times$ &   $100\%$ &  $60.5$ &  $94.9$ &  $89.3$ &  $87.6$ &  $72.1$ &  $86.7$ &  $85.9$ &  $91.1$ &  $70.1$ &  $80.4$ \\
Adapters ($8$-$256$)      &  $1.3\times$ &   $3.6\%$ &  $59.5$ &  $94.0$ &  $89.5$ &  $86.9$ &  $71.8$ &  $84.9$ &  $85.1$ &  $90.7$ &  $71.5$ &  $80.0$ \\
Adapters ($64$)           &  $1.2\times$ &   $2.1\%$ &  $56.9$ &  $94.2$ &  $89.6$ &  $87.3$ &  $71.8$ &  $85.3$ &  $84.6$ &  $91.4$ &  $68.8$ &  $79.6$ \\
\bottomrule
\end{tabular}
\end{adjustbox}
\caption{
Results on GLUE test sets scored using the GLUE evaluation server.
MRPC and QQP are evaluated using F1 score.
STS-B is evaluated using Spearman's correlation coefficient.
CoLA is evaluated using Matthew's Correlation.
The other tasks are evaluated using accuracy.
Adapter tuning achieves comparable overall score ($80.0$) to full fine-tuning ($80.4$) using $1.3\times$ parameters in total, compared to $9\times$.
Fixing the adapter size to $64$ leads to a slightly decreased overall score of $79.6$ and slightly smaller model.
\label{tab:glue}}
\end{table*}

We first evaluate on GLUE.\footnote{
We omit WNLI as in~\citet{devlin2018bert} because the no current algorithm beats the baseline of predicting the majority class.}
For these datasets, we transfer from the pre-trained BERT\textsubscript{LARGE} model,
which contains $24$ layers, and a total of $330$M parameters, see~\citet{devlin2018bert} for details.
We perform a small hyperparameter sweep for adapter tuning:
We sweep learning rates in $\{3 \cdot 10^{-5}, 3 \cdot 10^{-4}, 3 \cdot 10^{-3}\}$, and number of epochs in $\{3, 20\}$.
We test both using a fixed adapter size (number of units in the bottleneck),
and selecting the best size per task from $\{8, 64, 256\}$.
The adapter size is the only adapter-specific hyperparameter that we tune.
Finally, due to training instability, we re-run $5$ times with different random seeds and select the best model on the validation set.

Table~\ref{tab:glue} summarizes the results.
Adapters achieve a mean GLUE score of $80.0$, compared to $80.4$ achieved by full fine-tuning.
The optimal adapter size varies per dataset. For example, $256$ is chosen for MNLI, whereas for the smallest dataset, RTE, $8$ is chosen.
Restricting always to size $64$, leads to a small decrease in average accuracy to $79.6$.
To solve all of the datasets in Table~\ref{tab:glue}, fine-tuning requires $9\times$ the total number of BERT parameters.\footnote{
We treat MNLI\textsubscript{m} and MNLI\textsubscript{mm} as separate tasks with individually tuned hyperparameters.
However, they could be combined into one model, leaving $8\times$ overall.}
In contrast, adapters require only $1.3\times$ parameters.

\subsection{Additional Classification Tasks}

\renewcommand{\arraystretch}{0.9}
\begin{table*}[t]
\centering
\begin{adjustbox}{max width=\textwidth}
\begin{tabular}{l|rrrrr}
\toprule
Dataset &
\pbox{5cm}{No BERT\\baseline} &
\pbox{5cm}{BERT\textsubscript{BASE}\\Fine-tune} &
\pbox{5cm}{BERT\textsubscript{BASE}\\Variable FT} &
\pbox{5cm}{BERT\textsubscript{BASE}\\Adapters} \\
\midrule
20 newsgroups & $ 91.1 $ & $ 92.8 \pm 0.1 $ & $ 92.8 \pm 0.1 $ & $91.7 \pm 0.2$ \\
Crowdflower airline & $ 84.5 $ & $ 83.6 \pm 0.3 $ & $ 84.0 \pm 0.1 $ & $ 84.5 \pm 0.2 $ \\
Crowdflower corporate messaging & $ 91.9 $ & $ 92.5 \pm 0.5 $ & $ 92.4 \pm 0.6 $ & $ 92.9 \pm 0.3 $ \\
Crowdflower disasters & $ 84.9 $ & $ 85.3 \pm 0.4 $ & $ 85.3 \pm 0.4 $ & $ 84.1 \pm 0.2 $ \\
Crowdflower economic news relevance & $ 81.1 $ & $ 82.1 \pm 0.0 $ & $ 78.9 \pm 2.8 $ & $ 82.5 \pm 0.3 $ \\
Crowdflower emotion & $ 36.3 $ & $ 38.4 \pm 0.1 $ & $ 37.6 \pm 0.2 $ & $ 38.7 \pm 0.1 $ \\
Crowdflower global warming & $ 82.7 $ & $ 84.2 \pm 0.4
 $ & $ 81.9 \pm 0.2 $ & $ 82.7 \pm 0.3 $ \\
Crowdflower political audience & $ 81.0 $ & $ 80.9 \pm 0.3 $ & $ 80.7 \pm 0.8
 $ & $ 79.0 \pm 0.5 $ \\
Crowdflower political bias & $ 76.8 $ & $ 75.2 \pm 0.9 $ & $ 76.5 \pm 0.4 $ & $ 75.9 \pm 0.3 $ \\
Crowdflower political message & $ 43.8 $ & $ 38.9 \pm 0.6 $ & $ 44.9 \pm 0.6 $ & $ 44.1 \pm 0.2 $ \\
Crowdflower primary emotions & $ 33.5 $ & $ 36.9 \pm 1.6 $ & $ 38.2 \pm 1.0 $ & $ 33.9 \pm 1.4 $ \\
Crowdflower progressive opinion & $ 70.6 $ & $ 71.6 \pm 0.5 $ & $ 75.9 \pm 1.3 $ & $ 71.7 \pm 1.1 $ \\
Crowdflower progressive stance & $ 54.3 $ & $ 63.8 \pm 1.0 $ & $ 61.5 \pm 1.3 $ & $ 60.6 \pm 1.4 $ \\
Crowdflower US economic performance & $ 75.6 $ & $ 75.3 \pm 0.1 $ & $ 76.5 \pm 0.4 $ & $ 77.3 \pm 0.1 $ \\
Customer complaint database & $ 54.5 $ & $ 55.9 \pm 0.1 $ & $ 56.4 \pm 0.1 $ & $55.4 \pm 0.1$ \\
News aggregator dataset & $ 95.2 $ & $ 96.3 \pm 0.0 $ & $ 96.5 \pm 0.0 $ & $ 96.2 \pm 0.0 $ \\
SMS spam collection & $ 98.5 $ & $ 99.3 \pm 0.2 $ & $ 99.3 \pm 0.2 $ & $ 95.1 \pm 2.2 $ \\
\midrule
Average & $72.7$ & $73.7$ & $74.0$ & $73.3$  \\
\midrule
Total number of params & \textemdash & $17\times$ & $9.9\times$ & $1.19\times$   \\
Trained params/task & \textemdash & $100\%$ & $52.9\%$ & $1.14\%$ \\
\bottomrule
\end{tabular}
\end{adjustbox}
\caption{
Test accuracy for additional classification tasks.
In these experiments we transfer from the BERT\textsubscript{BASE} model.
For each task and algorithm, the model with the best validation set accuracy is chosen.
We report the mean test accuracy and s.e.m. across runs with different random seeds.}
\label{tab:hub_results}
\vskip-3mm
\end{table*}
\renewcommand{\arraystretch}{1.0}

To further validate that adapters yields compact, performant, models, we test on additional, publicly available, text classification tasks.
This suite contains a diverse set of tasks:
The number of training examples ranges from $900$ to $330$k,
the number of classes ranges from $2$ to $157$,
and the average text length ranging from $57$ to $1.9$k characters.
Statistics and references for all of the datasets are in the appendix.

For these datasets, we use a batch size of $32$.
The datasets are diverse, so we sweep a wide range of learning rates:
$\{1 \cdot 10^{-5}, 3 \cdot 10^{-5}, 1 \cdot 10^{-4}, 3 \cdot 10^{-3}\}$.
Due to the large number of datasets, we select the number of training epochs from the set $\{20, 50, 100\}$ manually, from inspection of the validation set learning curves.
We select the optimal values for both fine-tuning and adapters; the exact values are in the appendix.

We test adapters sizes in $\{2, 4, 8, 16, 32, 64\}$.
Since some of the datasets are small, fine-tuning the entire network may be sub-optimal.
Therefore, we run an additional baseline: variable fine-tuning.
For this, we fine-tune only the top $n$ layers, and freeze the remainder.
We sweep $n\in\{1,2,3,5,7,9,11,12\}$.
In these experiments, we use the BERT\textsubscript{BASE} model with $12$ layers,
therefore, variable fine-tuning subsumes full fine-tuning when $n=12$.

\begin{figure*}[t]
\centering
\vskip-1mm
\begin{tabular}{cc}
GLUE (BERT\textsubscript{LARGE}) & Additional Tasks (BERT\textsubscript{BASE}) \\
\includegraphics[width=0.45\textwidth]{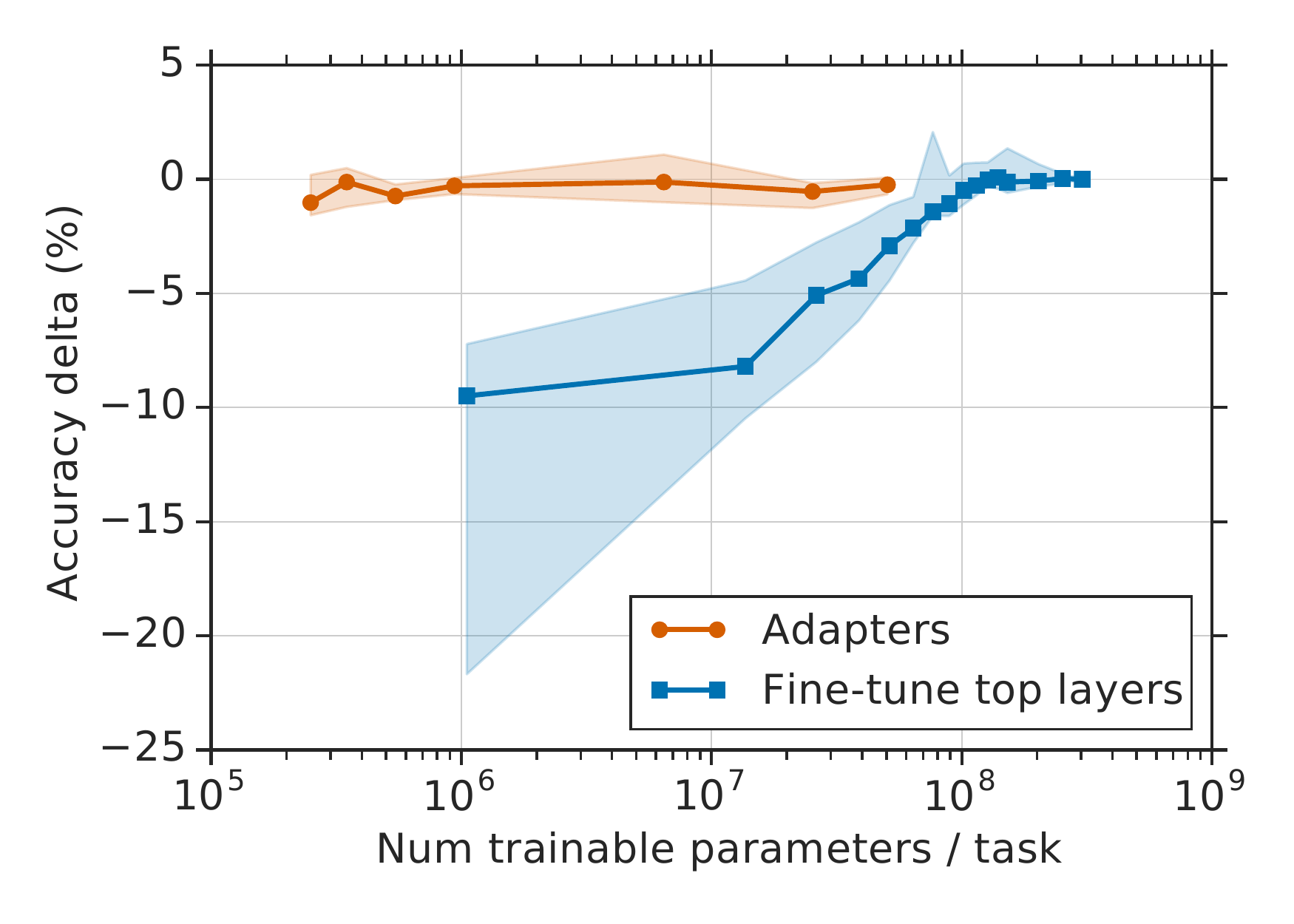}&
\includegraphics[width=0.45\textwidth]{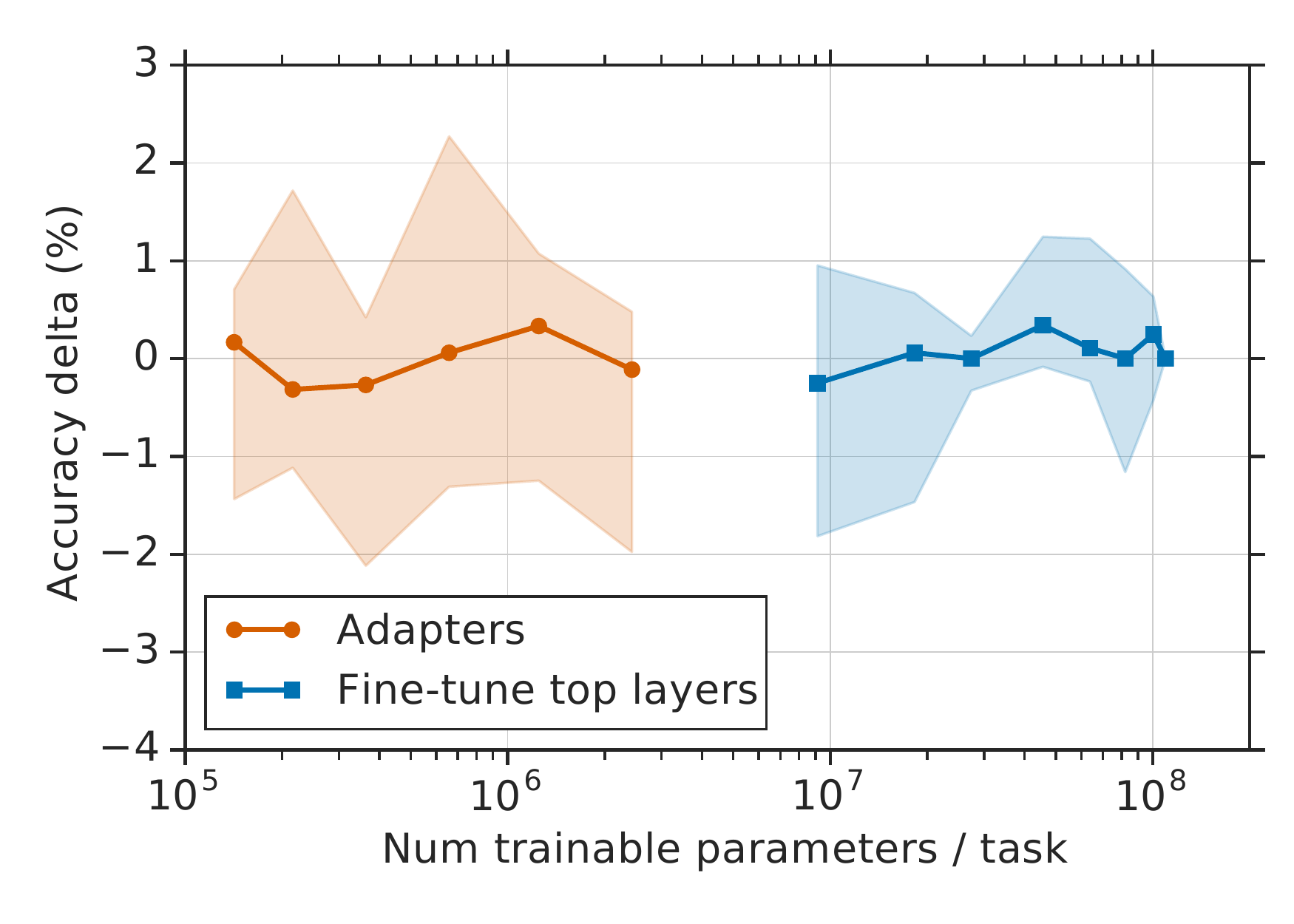}
\end{tabular}
\vskip-2mm
\caption{
Accuracy versus the number of trained parameters, aggregated across tasks.
We compare adapters of different sizes (orange) with fine-tuning the top $n$ layers, for varying $n$ (blue).
The lines and shaded areas indicate the $20$th, $50$th, and $80$th percentiles across tasks.
For each task and algorithm, the best model is selected for each point along the curve.
For GLUE, the validation set accuracy is reported.
For the additional tasks, we report the test-set accuracies.
To remove the intra-task variance in scores,
we normalize the scores for each model and task by subtracting the performance of full fine-tuning on the corresponding task.
}
\label{fig:tradeoff_alltasks}
\end{figure*}

\begin{figure*}[h!]
\centering
\vskip-1mm
\begin{tabular}{cc}
MNLI\textsubscript{m}(BERT\textsubscript{BASE}) & CoLA (BERT\textsubscript{BASE}) \\
\includegraphics[width=0.45\linewidth]{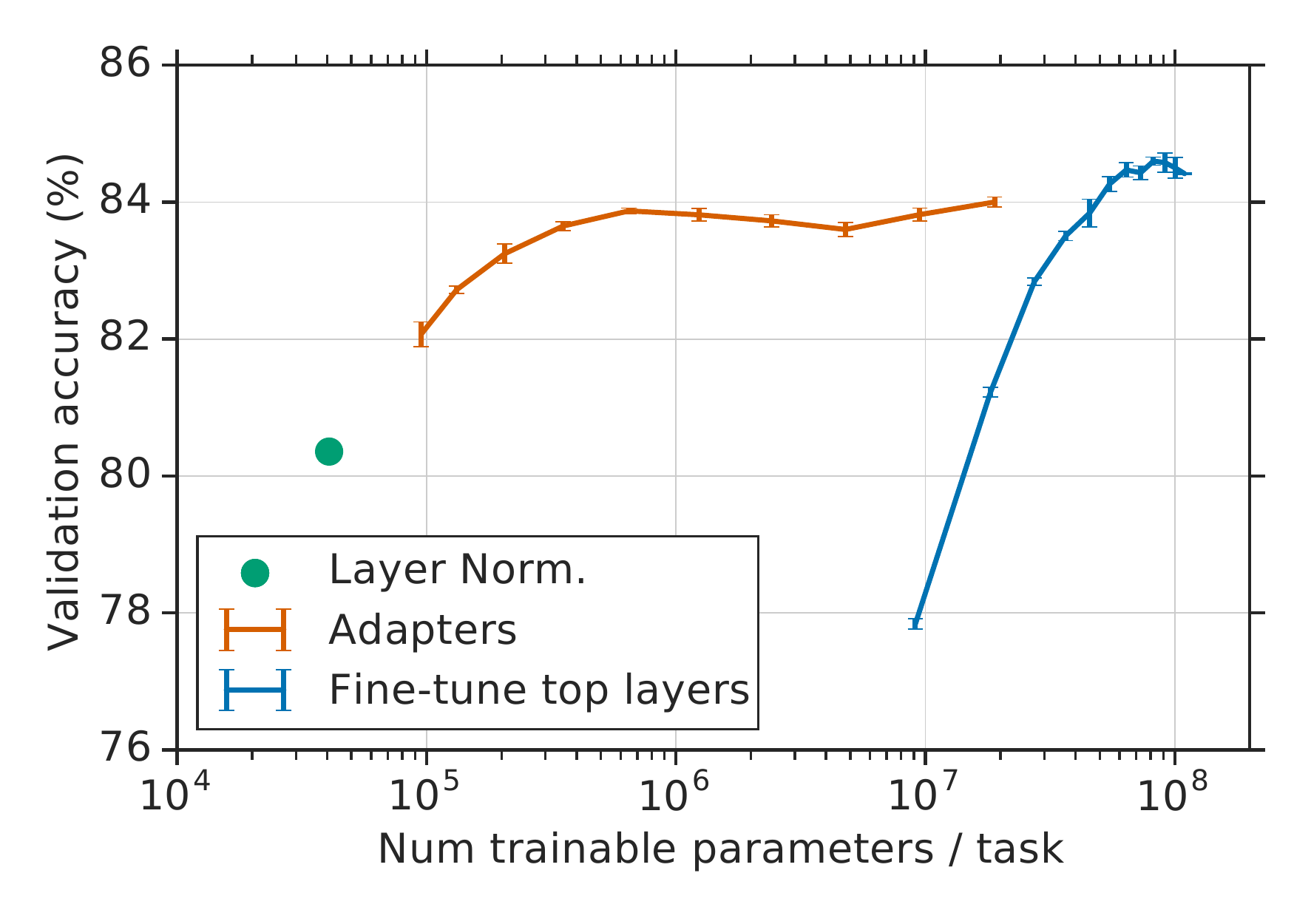}&
\includegraphics[width=0.45\linewidth]{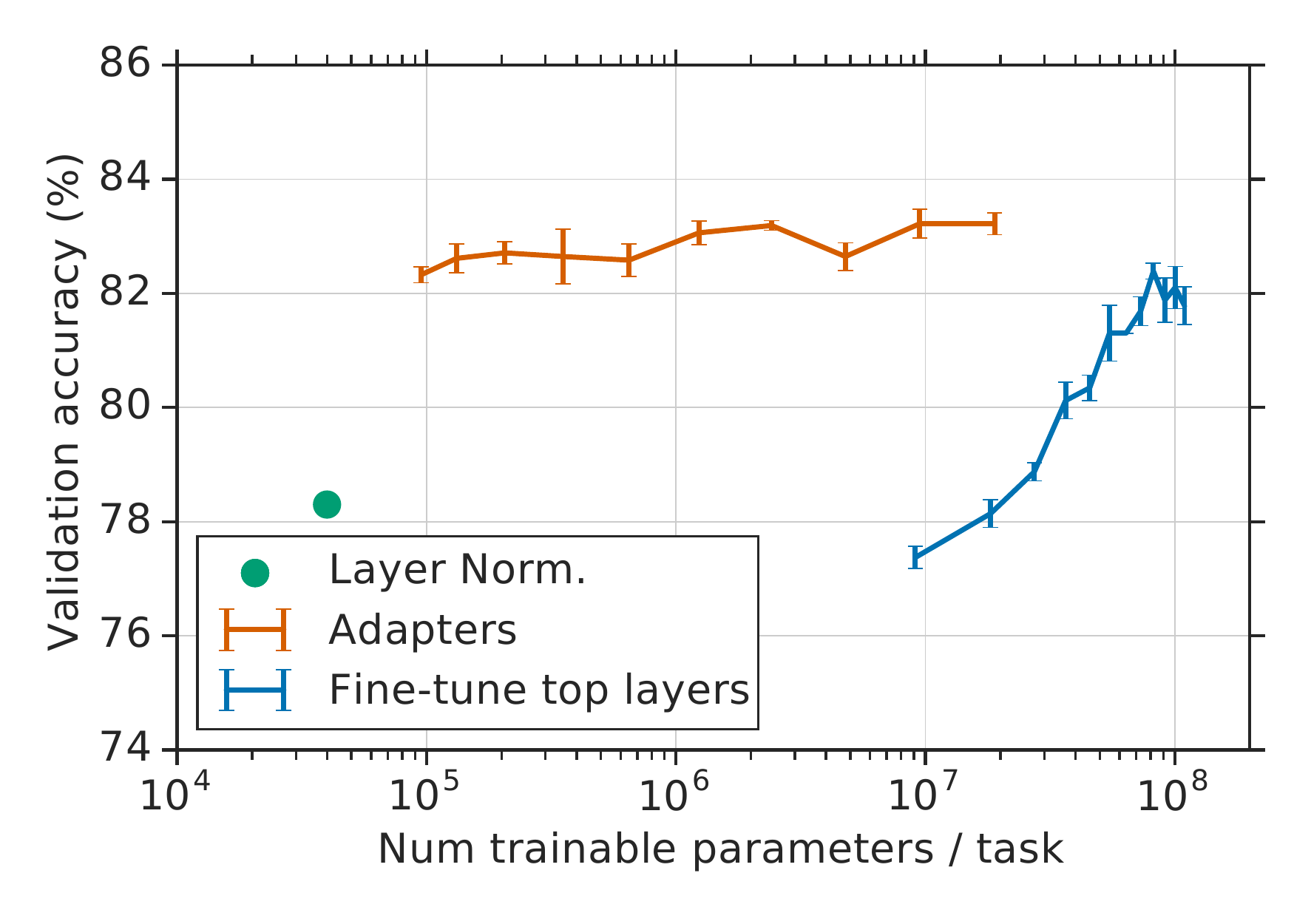}
\end{tabular}
\caption{
Validation set accuracy versus number of trained parameters for three methods:
(i) Adapter tuning with an adapter sizes $2^n$ for $n=0 \ldots 9$ (orange).
(ii) Fine-tuning the top $k$ layers for $k=1\ldots 12$ (blue).
(iii) Tuning the layer normalization parameters only (green).
Error bars indicate $\pm 1$ s.e.m. across three random seeds.}
\label{fig:tradeoff_glue}
\vskip-3mm
\end{figure*}

Unlike the GLUE tasks, there is no comprehensive set of state-of-the-art numbers for this suite of tasks.
Therefore, to confirm that our BERT-based models are competitive, we collect our own benchmark performances.
For this, we run a large-scale hyperparameter search over standard network topologies.
Specifically, we run the single-task Neural AutoML algorithm, similar to~\citet{zoph2017,wong2018transferautoml}.
This algorithm searches over a space of feedforward and convolutional networks,
stacked on pre-trained text embeddings modules publicly available via TensorFlow Hub\footnote{\url{https://www.tensorflow.org/hub}}.
The embeddings coming from the TensorFlow Hub modules may be frozen or fine-tuned.
The full search space is described in the appendix.
For each task, we run AutoML for one week on CPUs, using $30$ machines.
In this time the algorithm explores over $10$k models on average per task.
We select the best final model for each task according to validation set accuracy.

The results for the AutoML benchmark (``no BERT baseline''), fine-tuning, variable fine-tuning, and adapter-tuning are reported in Table~\ref{tab:hub_results}.
The AutoML baseline demonstrates that the BERT models are competitive.
This baseline explores thousands of models, yet the BERT models perform better on average.
We see similar pattern of results to GLUE.
The performance of adapter-tuning is close to full fine-tuning ($0.4\%$ behind).
Fine-tuning requires $17\times$ the number of parameters to BERT\textsubscript{BASE} to solve all tasks.
Variable fine-tuning performs slightly better than fine-tuning, whilst training fewer layers.
The optimal setting of variable fine-tuning results in training $52\%$ of the network on average per task, reducing the total to $9.9\times$ parameters.
Adapters, however, offer a much more compact model.
They introduce $1.14\%$ new parameters per task, resulting in $1.19\times$ parameters for all $17$ tasks.

\subsection{Parameter/Performance trade-off\label{sec:param_efficiency}}

The adapter size controls the parameter efficiency, smaller adapters introduce fewer parameters, at a possible cost to performance.
To explore this trade-off, we consider different adapter sizes, and compare to two baselines:
(i) Fine-tuning of only the top $k$ layers of BERT\textsubscript{BASE}.
(ii) Tuning only the layer normalization parameters.
The learning rate is tuned using the range presented in Section~\ref{sec:glue}.

Figure~\ref{fig:tradeoff_alltasks} shows the parameter/performance trade-off aggregated over all classification tasks in each suite (GLUE and ``additional'').
On GLUE, performance decreases dramatically when fewer layers are fine-tuned.
Some of the additional tasks benefit from training fewer layers, so performance of fine-tuning decays much less.
In both cases, adapters yield good performance across a range of sizes two orders of magnitude fewer than fine-tuning.

Figure~\ref{fig:tradeoff_glue} shows more details for two GLUE tasks: MNLI\textsubscript{m} and CoLA.
Tuning the top layers trains more task-specific parameters for all $k>2$.
When fine-tuning using a comparable number of task-specific parameters, the performance decreases substantially compared to adapters.
For instance, fine-tuning just the top layer yields approximately $9$M trainable parameters and $77.8 \% \pm 0.1 \%$ validation accuracy on MNLI\textsubscript{m}.
In contrast, adapter tuning with size $64$ yields approximately $2$M trainable parameters and $83.7\% \pm 0.1 \%$  validation accuracy.
For comparison, full fine-tuning attains $84.4 \% \pm 0.02 \%$ on MNLI\textsubscript{m}.
We observe a similar trend on CoLA.

As a further comparison, we tune the parameters of layer normalization alone.
These layers only contain point-wise additions and multiplications, so introduce very few trainable parameters: $40$k for BERT\textsubscript{BASE}.
However this strategy performs poorly: performance decreases by approximately $3.5\%$ on CoLA and $4\%$ on MNLI.

To summarize, adapter tuning is highly parameter-efficient, and produces a compact model with a strong performance, comparable to full fine-tuning.
Training adapters with sizes $0.5-5\%$ of the original model,
performance is within $1\%$ of the competitive published results on BERT\textsubscript{LARGE}.

\subsection{SQuAD Extractive Question Answering}

\begin{figure}[t]
\centering
\vskip-2mm
\includegraphics[width=0.9\linewidth]{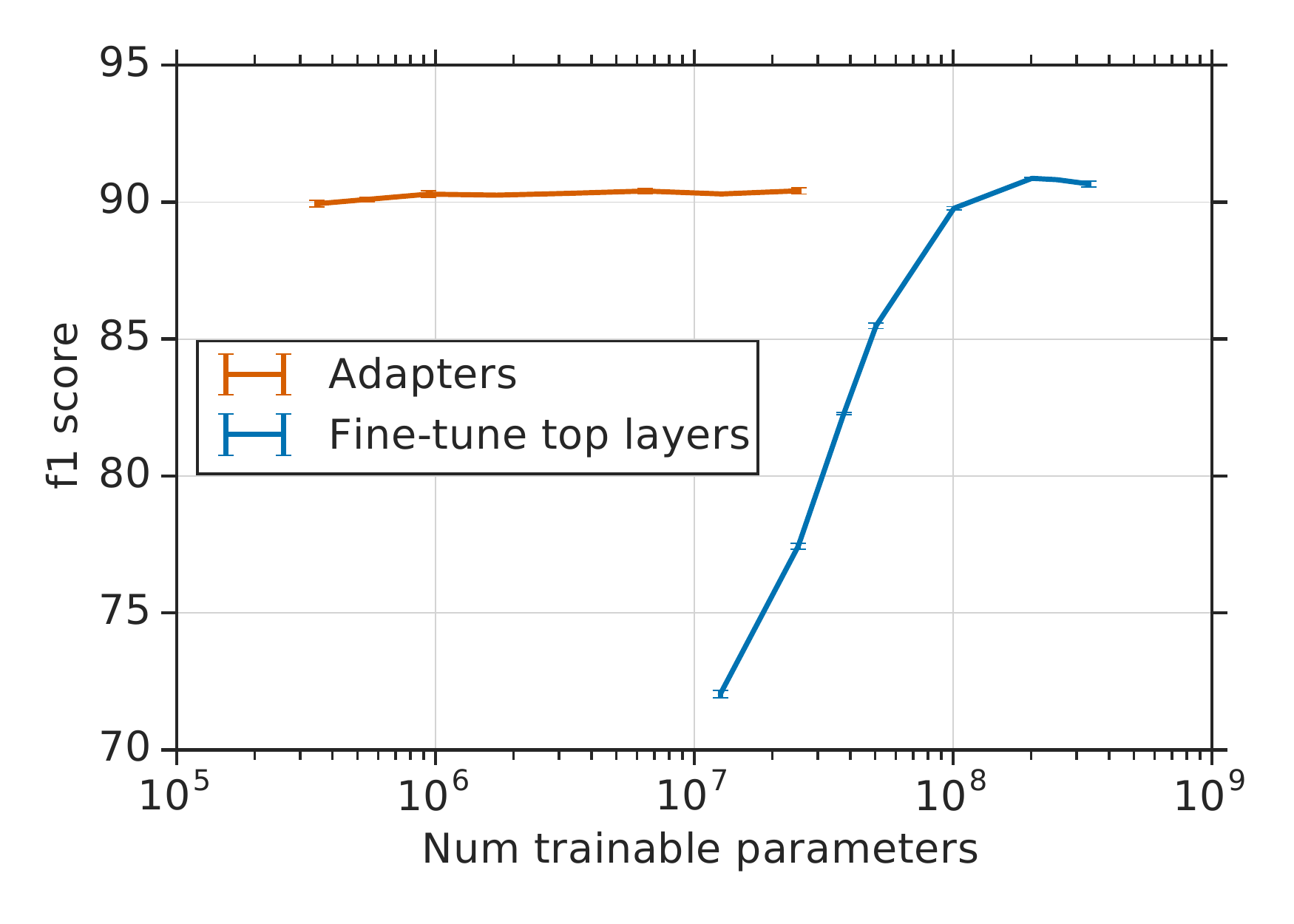}
\caption{
Validation accuracy versus the number of trained parameters for SQuAD v1.1.
Error bars indicate the s.e.m. across three seeds, using the best hyperparameters.
}
\label{fig:squad}
\vskip-5mm
\end{figure}

Finally, we confirm that adapters work on tasks other than classification by running on SQuAD v1.1~\citep{rajpurkar2018}.
Given a question and Wikipedia paragraph, this task requires selecting the answer span to the question from the paragraph.
Figure~\ref{fig:squad} displays the parameter/performance trade-off of fine-tuning and adapters on the SQuAD validation set.
For fine-tuning, we sweep the number of trained layers, learning rate in $\{3\cdot 10^{-5}, 5\cdot 10^{-5}, 1\cdot 10^{-4}\}$, and number of epochs in $\{2,3,5\}$.
For adapters, we sweep the adapter size, learning rate in $\{3\cdot 10^{-5}, 1\cdot 10^{-4}, 3\cdot 10^{-4}, 1\cdot 10^{-3}\}$, and number of epochs in $\{3,10,20\}$.
As for classification, adapters attain performance comparable to full fine-tuning, while training many fewer parameters.
Adapters of size $64$ ($2\%$ parameters) attain a best F1 of $90.4$\%, while fine-tuning attains $90.7$.
SQuAD performs well even with very small adapters, those of size $2$ ($0.1\%$ parameters) attain an F1 of $89.9$.

\subsection{Analysis and Discussion\label{sec:discussion}}

\begin{figure*}
\centering
\begin{tabular}[t]{ccc}
MNLI\textsubscript{m} & CoLA & \\
\includegraphics[height=3.7cm]{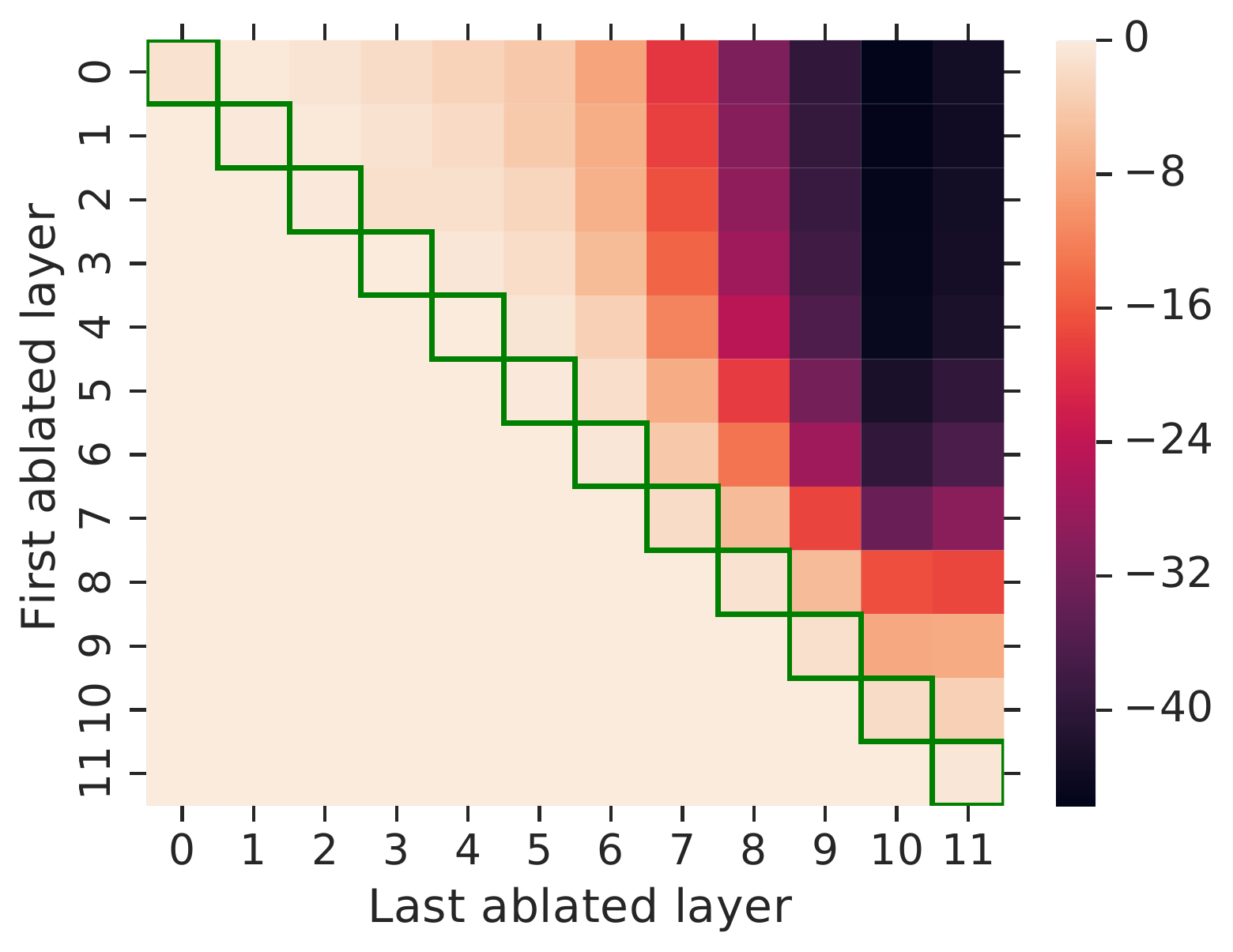}&
\includegraphics[height=3.7cm]{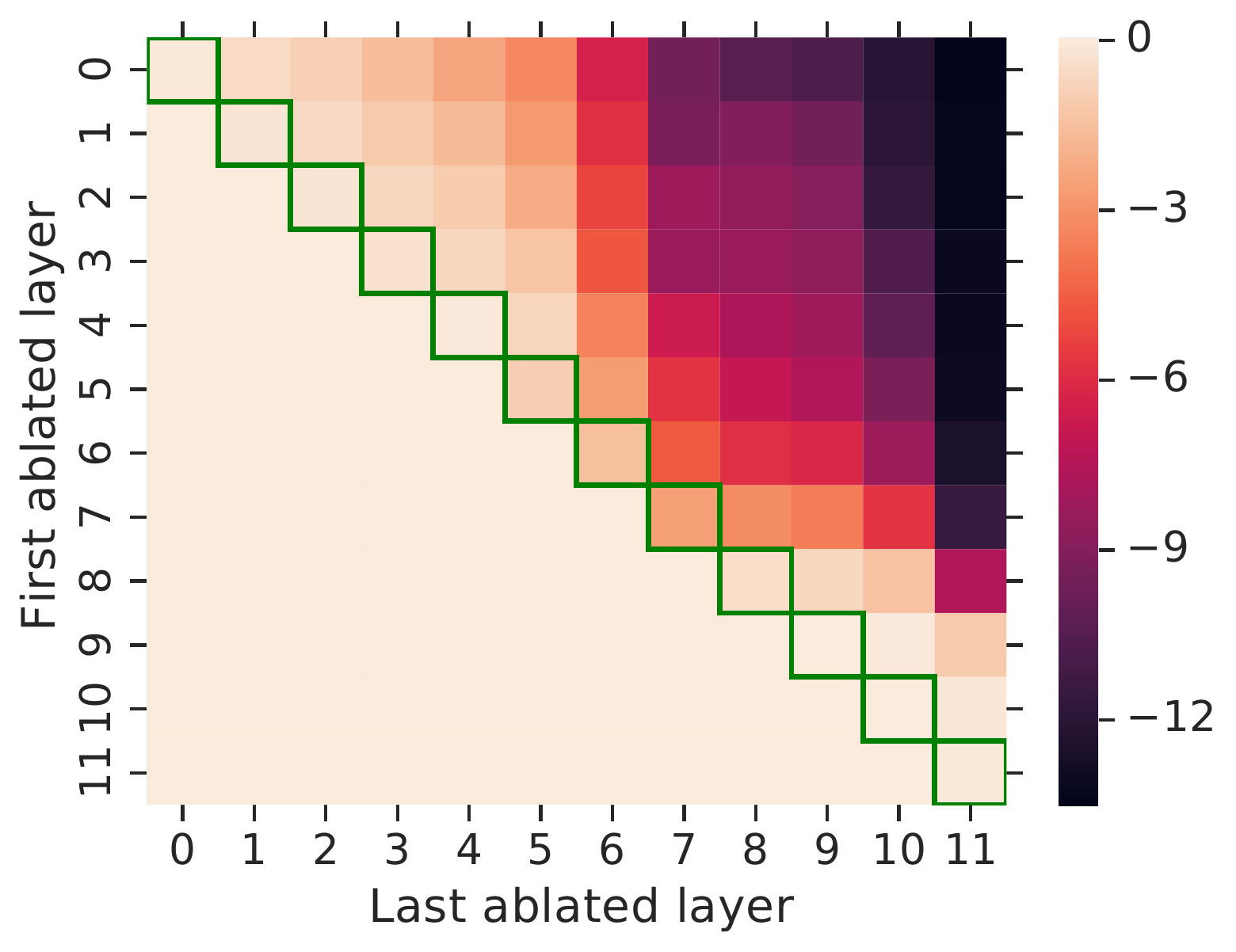}&
\raisebox{-0.2cm}{\includegraphics[height=3.9cm]{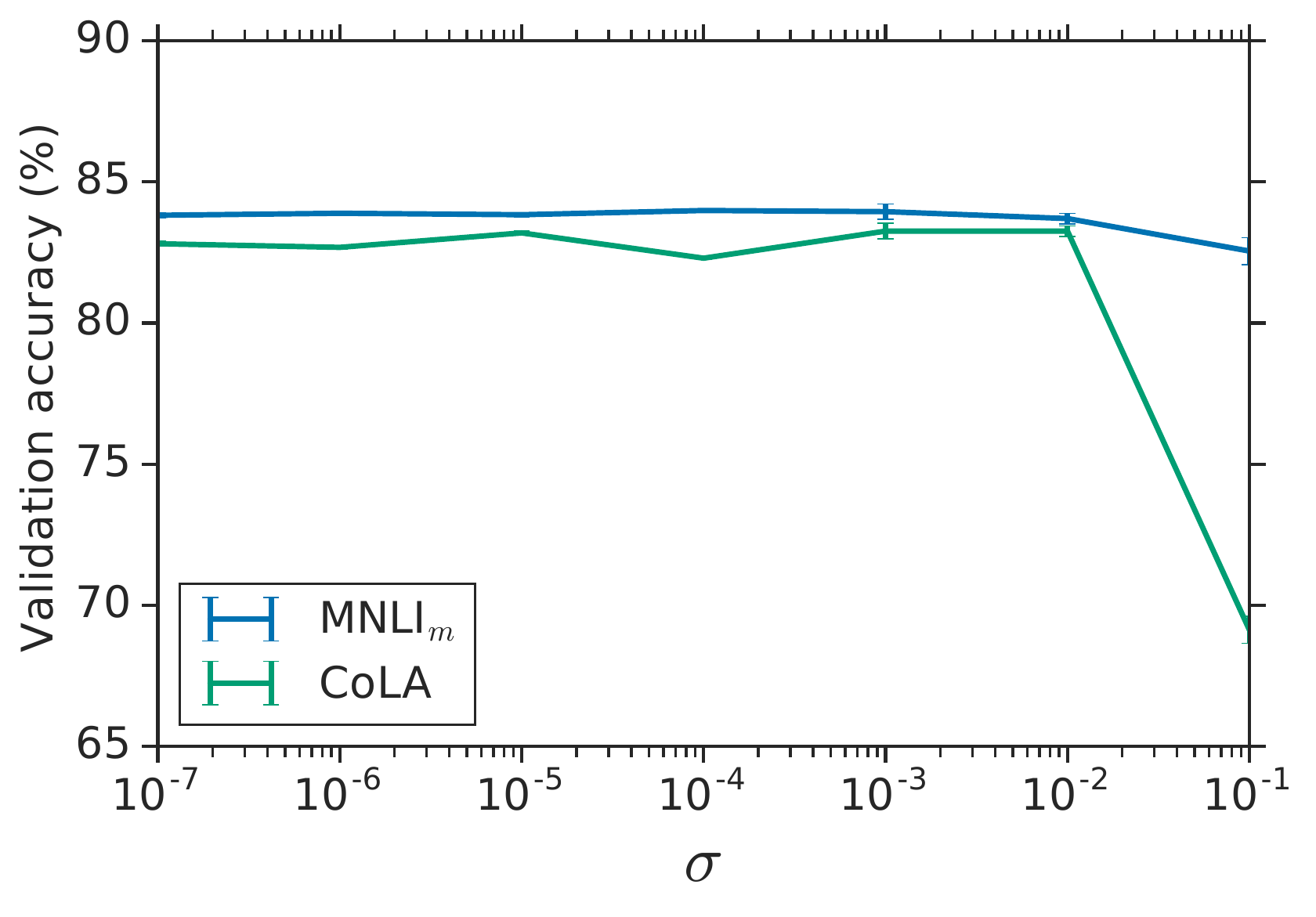}}
\end{tabular}
\caption{
\textbf{Left, Center:}
Ablation of trained adapters from continuous layer spans.
The heatmap shows the relative decrease in validation accuracy to the fully trained adapted model.
The \emph{y} and \emph{x} axes indicate the first and last layers ablated (inclusive), respectively.
The diagonal cells, highlighted in green, indicate ablation of a single layer's adapters.
The cell in the top-right indicates ablation of all adapters.
Cells in the lower triangle are meaningless, and are set to $0\%$, the best possible relative performance.
\textbf{Right:}
Performance of BERT\textsubscript{BASE} using adapters with different initial weight magnitudes.
The \emph{x}-axis is the standard deviation of the initialization distribution.
}
\label{fig:ablation_and_init}
\vskip-4mm
\end{figure*}

We perform an ablation to determine which adapters are influential.
For this, we remove some trained adapters and re-evaluate the model (without re-training) on the validation set.
Figure~\ref{fig:ablation_and_init} shows the change in performance when removing adapters from all continuous layer spans.
The experiment is performed on BERT\textsubscript{BASE} with adapter size $64$ on MNLI and CoLA.

First, we observe that removing any single layer's adapters has only a small impact on performance.
The elements on the heatmaps' diagonals show the performances of removing adapters from single layers, where largest performance drop is $2\%$.
In contrast, when all of the adapters are removed from the network,
the performance drops substantially:
to $37\%$ on MNLI and $69\%$ on CoLA -- scores attained by predicting the majority class.
This indicates that although each adapter has a small influence on the overall network, the overall effect is large.

Second, Figure~\ref{fig:ablation_and_init} suggests that adapters on the lower layers have a smaller impact than the higher-layers.
Removing the adapters from the layers $0-4$ on MNLI barely affects performance.
This indicates that adapters perform well because they automatically prioritize higher layers.
Indeed, focusing on the upper layers is a popular strategy in fine-tuning~\citep{howard2018universal}.
One intuition is that the lower layers extract lower-level features that are shared among tasks, while the
higher layers build features that are unique to different tasks.
This relates to our observation that for some tasks, fine-tuning only the top layers outperforms full fine-tuning, see Table~\ref{tab:hub_results}.

Next, we investigate the robustness of the adapter modules to the number of neurons and initialization scale.
In our main experiments the weights in the adapter module were drawn from
a zero-mean Gaussian with standard deviation $10^{-2}$, truncated to two standard deviations.
To analyze the impact of the initialization scale on the performance, we test standard deviations in the interval $[10^{-7},1]$.
Figure~\ref{fig:ablation_and_init} summarizes the results.
We observe that on both datasets, the performance of adapters is robust for standard deviations below $10^{-2}$.
However, when the initialization is too large, performance degrades, more substantially on CoLA.

To investigate robustness of adapters to the number of neurons, we re-examine the experimental data from Section~\ref{sec:glue}.
We find that the quality of the model across adapter sizes is stable,
and a fixed adapter size across all the tasks could be used with small detriment to performance.
For each adapter size we calculate the mean validation accuracy across the eight
classification tasks by selecting the optimal learning rate and number of epochs\footnote{
We treat here MNLI\textsubscript{m} and MNLI\textsubscript{mm} as separate tasks.
For consistency, for all datasets we use accuracy metric and exclude the regression STS-B task.}.
For adapter sizes $8$, $64$, and $256$, the mean validation accuracies are $86.2\%$, $85.8\%$ and $85.7\%$, respectively.
This message is further corroborated by Figures~\ref{fig:tradeoff_glue} and~\ref{fig:squad},
which show a stable performance across a few orders of magnitude.

Finally, we tried a number of extensions to the adapter's architecture
that did not yield a significant boost in performance.
We document them here for completeness.
We experimented with
(i) adding a batch/layer normalization to the adapter,
(ii) increasing the number of layers per adapter,
(iii) different activation functions, such as tanh,
(iv) inserting adapters only inside the attention layer,
(v) adding adapters in parallel to the main layers, and possibly with a multiplicative interaction.
In all cases we observed the resulting performance to be similar to the bottleneck proposed in Section~\ref{sec:bottleneckadapter}.
Therefore, due to its simplicity and strong performance, we recommend the original adapter architecture.

\section{Related Work}

\paragraph{Pre-trained text representations}
Pre-trained textual representations are widely used to improve performance on NLP tasks.
These representations are trained on large corpora (usually unsupervised), and fed as features to downstream models.
In deep networks, these features may also be fine-tuned on the downstream task.
Brown clusters, trained on distributional information, are a classic example of pre-trained representations~\citep{brown1992}.
\citet{turian2010} show that pre-trained embeddings of words outperform those trained from scratch.
Since deep-learning became popular, word embeddings have been widely used, and many training strategies have arisen~\citep{mikolov2013,pennington2014,bojanowski2017enriching}.
Embeddings of longer texts, sentences and paragraphs, have also been developed~\citep{le2014,kiros2015,conneau2017,cer2019}.

To encode context in these representations, features are extracted from internal representations of sequence models,
such as MT systems~\citep{mccann2017}, and BiLSTM language models, as used in ELMo~\citep{peters2018}.
As with adapters, ELMo exploits the layers other than the top layer of a pre-trained network.
However, this strategy only \emph{reads} from the inner layers.
In contrast, adapters \emph{write} to the inner layers, re-configuring the processing of features through the entire network.

\paragraph{Fine-tuning}
Fine-tuning an entire pre-trained model has become a popular alternative to features~\citep{dai2015,howard2018universal,radford2018improving}
In NLP, the upstream model is usually a neural language model~\citep{bengio2003}.
Recent state-of-the-art results on question answering~\citep{rajpurkar2016} and text classification~\citep{wang2018glue} have been attained by fine-tuning a Transformer network~\citep{vaswani2017} with a Masked Language Model loss~\citep{devlin2018bert}.
Performance aside, an advantage of fine-tuning is that it does not require task-specific model design, unlike representation-based transfer.
However, vanilla fine-tuning does require a new set of network weights for every new task.

\paragraph{Multi-task Learning}
Multi-task learning (MTL) involves training on tasks simultaneously.
Early work shows that sharing network parameters across tasks exploits task regularities, yielding improved performance~\citep{caruana1997}.
The authors share weights in lower layers of a network, and use specialized higher layers.
Many NLP systems have exploited MTL.
Some examples include: text processing systems (part of speech, chunking, named entity recognition, etc.)~\citep{collobert2008}, multilingual models~\citep{huang2013cross}, semantic parsing~\citep{peng2017}, machine translation~\citep{johnson2017}, and question answering~\citep{choi2017}.
MTL yields a single model to solve all problems.
However, unlike our adapters, MTL requires simultaneous access to the tasks during training.

\paragraph{Continual Learning}
As an alternative to simultaneous training, continual, or lifelong, learning aims to learn from a sequence of tasks~\citep{thrun1998}.
However, when re-trained, deep networks tend to forget how to perform previous tasks; a challenge termed catastrophic forgetting~\citep{mccloskey1989catastrophic,french1999catastrophic}.
Techniques have been proposed to mitigate forgetting~\citep{kirkpatrick2017overcoming,zenke2017continual}, however, unlike for adapters, the memory is imperfect.
Progressive Networks avoid forgetting by instantiating a new network ``column'' for each task~\citep{rusu2016progressive}.
However, the number of parameters grows linearly with the number of tasks,
since adapters are very small, our models scale much more favorably.

\paragraph{Transfer Learning in Vision}
Fine-tuning models pre-trained on ImageNet~\citep{deng2009} is ubiquitous when building image recognition models~\citep{yosinski2014,huh2016makes}.
This technique attains state-of-the-art performance on many vision tasks, including classification~\citep{kornblith2018better}, fine-grained classifcation~\citep{hermans2017}, segmentation~\citep{long2015}, and detection~\citep{girshick2014}.
In vision, convolutional adapter modules have been studied~\citep{rebuffi2017,rebuffi2018,rosenfeld2018incremental}.
These works perform incremental learning in multiple domains by adding small convolutional layers to a ResNet~\citep{he2016} or VGG net~\citep{simonyan2014very}.
Adapter size is limited using $1\times 1$ convolutions, whilst the original networks typically use $3\times 3$.
This yields $11\%$ increase in overall model size per task.
Since the kernel size cannot be further reduced other weight compression techniques must be used to attain further savings.
Our bottleneck adapters can be much smaller, and still perform well.

Concurrent work explores similar ideas for BERT~\citep{stickland2019bert}.
The authors introduce Projected Attention Layers (PALs), small layers with a similar role to our adapters.
The main differences are i) \citet{stickland2019bert} use a different architecture,
and ii) they perform multitask training, jointly fine-tuning BERT on all GLUE tasks.
\citet{semnani2019} perform an emprical comparison of our bottleneck Adpaters and PALs on SQuAD v2.0~\citep{rajpurkar2018}.

\subsubsection*{Acknowledgments}
We would like to thank Andrey Khorlin, Lucas Beyer,
No\'e Lutz, and Jeremiah Harmsen for useful comments and discussions.

\bibliography{nlp}
\bibliographystyle{icml2019}

\clearpage

\appendix
\onecolumn

\icmltitle{Supplementary Material for\\Parameter-Efficient Transfer Learning for NLP}

\section{Additional Text Classification Tasks}
\label{appendix:hub_stats}

\begin{table*}[ht]
\centering
\begin{adjustbox}{max width=\textwidth}
\begin{tabular}{l|rrrrrr}
\toprule
Dataset & Train examples & Validation examples & Test examples & Classes & Avg text length & Reference \\
\midrule
20 newsgroups & $15076$ & $1885$ & $1885$ & $20$ & $1903$ & \citep{Lang95} \\
Crowdflower airline & $11712$ & $1464$ & $1464$ & $3$ & $104$ & crowdflower.com \\
Crowdflower corporate messaging & $2494$ & $312$ & $312$ & $4$ & $121$ & crowdflower.com \\
Crowdflower disasters & $8688$ & $1086$ & $1086$ & $2$ & $101$ & crowdflower.com \\
Crowdflower economic news relevance & $6392$ & $799$ & $800$ & $2$ & $1400$ & crowdflower.com \\
Crowdflower emotion & $32000$ & $4000$ & $4000$ & $13$ & $73$ & crowdflower.com \\
Crowdflower global warming & $3380$ & $422$ & $423$ & $2$ & $112$ & crowdflower.com \\
Crowdflower political audience & $4000$ & $500$ & $500$ & $2$ & $205$ & crowdflower.com \\
Crowdflower political bias & $4000$ & $500$ & $500$ & $2$ & $205$ & crowdflower.com \\
Crowdflower political message & $4000$ & $500$ & $500$ & $9$ & $205$ & crowdflower.com \\
Crowdflower primary emotions & $2019$ & $252$ & $253$ & $18$ & $87$ & crowdflower.com \\
Crowdflower progressive opinion & $927$ & $116$ & $116$ & $3$ & $102$ & crowdflower.com \\
Crowdflower progressive stance & $927$ & $116$ & $116$ & $4$ & $102$ & crowdflower.com \\
Crowdflower US economic performance & $3961$ & $495$ & $496$ & $2$ & $305$ & crowdflower.com \\
Customer complaint database & $146667$ & $18333$ & $18334$ & $157$ & $1046$ & catalog.data.gov \\
News aggregator dataset & $338349$ & $42294$ & $42294$ & $4$ & $57$ & \citep{Lichman:2013}\\
SMS spam collection & $4459$ & $557$ & $558$ & $2$ & $81$ & \citep{Almeida:2011:CSS:2034691.2034742}\\
\bottomrule
\end{tabular}
\end{adjustbox}
\caption{Statistics and references for the additional text classification tasks.}
\label{tab:hub_stats}
\end{table*}

\begin{table*}[ht]
\centering
\begin{adjustbox}{max width=\textwidth}
\begin{tabular}{l|rr}
\toprule
Dataset & Epochs (Fine-tune) & Epochs (Adapters) \\
\midrule
20 newsgroups & $50$ & $50$ \\
Crowdflower airline & $50$ & $20$ \\
Crowdflower corporate messaging & $100$ & $50$ \\
Crowdflower disasters & $50$ & $50$ \\
Crowdflower economic news relevance & $20$ & $20$ \\
Crowdflower emotion & $20$ & $20$ \\
Crowdflower global warming & $100$ & $50$ \\
Crowdflower political audience & $50$ & $20$ \\
Crowdflower political bias & $50$ & $50$ \\
Crowdflower political message & $50$ & $50$ \\
Crowdflower primary emotions & $100$ & $100$ \\
Crowdflower progressive opinion & $100$ & $100$ \\
Crowdflower progressive stance & $100$ & $100$ \\
Crowdflower US economic performance & $100$ & $20$ \\
Customer complaint database & $20$ & $20$ \\
News aggregator dataset & $20$ & $20$ \\
SMS spam collection & $50$ & $20$ \\
\bottomrule
\end{tabular}
\end{adjustbox}
\caption{Number of training epochs selected for the additional classification tasks.}
\label{tab:hub_epochs}
\end{table*}

\begin{table*}[ht]
\centering
\begin{adjustbox}{max width=\textwidth}
\begin{tabular}{ll}
\toprule
\bf Parameter  & \bf Search Space \\
\midrule
1) Input embedding modules & Refer to Table~\ref{tab:hub_embeddings_text} \\
2) Fine-tune input embedding module & \{True, False\} \\
3) Lowercase text & \{True, False\} \\
4) Remove non alphanumeric text & \{True, False\} \\
5) Use convolution & \{True, False\} \\
6) Convolution activation & \{relu, relu6, leaky relu, swish, sigmoid, tanh\} \\
7) Convolution batch norm & \{True, False\} \\
8) Convolution max ngram length & \{2, 3\} \\
9) Convolution dropout rate & [0.0, 0.4] \\
10) Convolution number of filters & [50, 200]\\
11) Convolution embedding dropout rate & [0.0, 0.4] \\
12) Number of hidden layers & \{0, 1, 2, 3, 5\} \\
13) Hidden layers size & \{64, 128, 256\} \\
14) Hidden layers activation & \{relu, relu6, leaky relu, swish, sigmoid, tanh\} \\
15) Hidden layers normalization & \{none, batch norm, layer norm\} \\
16) Hidden layers dropout rate & \{0.0, 0.05, 0.1, 0.2, 0.3, 0.4, 0.5\} \\
17) Deep tower learning rate & \{0.001, 0.005, 0.01, 0.05, 0.1, 0.5\} \\
18) Deep tower regularization weight & \{0.0, 0.0001, 0.001, 0.01\} \\
19) Wide tower learning rate & \{0.001, 0.005, 0.01, 0.05, 0.1, 0.5\} \\
20) Wide tower regularization weight & \{0.0, 0.0001, 0.001, 0.01\} \\
21) Number of training samples & \{1e5, 2e5, 5e5, 1e6, 2e6\} \\
\bottomrule
\end{tabular}
\end{adjustbox}
\caption{The search space of baseline models for the additional text classification tasks.}
\label{tab:hub_ss}
\end{table*}

\begin{table*}[ht]
\centering
\begin{adjustbox}{max width=\textwidth}
\begin{tabular}{llllll}
\bf ID & \bf Dataset &\bf Embed & \bf Vocab. & \bf Training & \bf
TensorFlow Hub Handles \\
\bf  & \bf  size & \bf dim. & \bf size & \bf algorithm &
Prefix: \texttt{https://tfhub.dev/google/}\\
& (tokens) & & & \\
\hline \\
English-small  & 7B & 50 & 982k & Lang. model &
\texttt{nnlm-en-dim50-with-normalization/1} \\
English-big & 200B & 128 & 999k & Lang. model &
\texttt{nnlm-en-dim128-with-normalization/1} \\
English-wiki-small & 4B & 250 & 1M & Skipgram &
\texttt{Wiki-words-250-with-normalization/1} \\
English-wiki-big & 4B & 500 & 1M & Skipgram &
\texttt{Wiki-words-500-with-normalization/1} \\
Universal-sentence-encoder & - & 512 & - & \citep{cer2018universal} &
\texttt{universal-sentence-encoder/2} \\
\end{tabular}
\end{adjustbox}
\caption{Options for text input embedding modules.
These are pre-trained text embedding tables.
We provide the handle for the modules that are publicly distributed via the TensorFlow Hub service (\texttt{https://www.tensorflow.org/hub}).}
\label{tab:hub_embeddings_text}
\end{table*}

\begin{table*}[ht]
\centering
\begin{adjustbox}{max width=\textwidth}
\begin{tabular}{l|lllllllllllllllllllll}
\toprule
Dataset & 1 & 2 & 3 & 4 & 5 & 6 & 7 & 8 & 9 & 10 & 11 & 12 & 13 & 14 & 15 & 16 & 17 & 18 & 19 & 20 & 21\\
\midrule
20 newsgroups & Universal-sentence-encoder & False & True & True & False & relu6 & False & 2 & 0.37 & 94 & 0.38 & 1 & 128 & leaky relu & batch norm & 0.5 & 0.5 & 0 & 0.05 & 0.0001 & 1000000 \\
Crowdflower airline & English-big & False & False & False & True & leaky relu & False & 3 & 0.36 & 200 & 0.07 & 0 & 128 & tanh & layer norm & 0.4 & 0.1 & 0.001 & 0.05 & 0.001 & 200000 \\
Crowdflower corporate messaging & English-big & False & False & True & True & tanh & True & 3 & 0.40 & 56 & 0.40 & 1 & 64 & tanh & batch norm & 0.5 & 0.5 & 0.001 & 0.01 & 0 & 200000 \\
Crowdflower disasters & Universal-sentence-encoder & True & True & False & True & swish & True & 3 & 0.27 & 52 & 0.22 & 0 & 64 & relu & none & 0.2 & 0.005 & 0.0001 & 0.005 & 0.01 & 500000 \\
Crowdflower economic news relevance & Universal-sentence-encoder & True & True & False & False & leaky relu & False & 2 & 0.27 & 63 & 0.04 & 3 & 128 & swish & layer norm & 0.2 & 0.01 & 0.01 & 0.001 & 0 & 100000 \\
Crowdflower emotion & Universal-sentence-encoder & False & True & False & False & relu6 & False & 3 & 0.35 & 132 & 0.34 & 1 & 64 & tanh & none & 0.05 & 0.05 & 0 & 0.05 & 0 & 200000 \\
Crowdflower global warming & Universal-sentence-encoder & False & True & True & False & swish & False & 3 & 0.39 & 200 & 0.36 & 1 & 128 & leaky relu & batch norm & 0.4 & 0.05 & 0 & 0.001 & 0.001 & 1000000 \\
Crowdflower political audience & English-small & True & False & True & True & relu & False & 3 & 0.11 & 98 & 0.07 & 0 & 64 & relu & none & 0.5 & 0.05 & 0.001 & 0.001 & 0 & 100000 \\
Crowdflower political bias & English-big & False & True & True & False & swish & False & 3 & 0.12 & 81 & 0.30 & 0 & 64 & relu6 & none & 0 & 0.01 & 0 & 0.005 & 0.01 & 200000 \\
Crowdflower political message & Universal-sentence-encoder & False & False & True & False & swish & True & 2 & 0.36 & 57 & 0.35 & 0 & 64 & tanh & none & 0.5 & 0.01 & 0.001 & 0.005 & 0 & 200000 \\
Crowdflower primary emotions & English-big & False & True & True & True & swish & False & 3 & 0.40 & 191 & 0.03 & 0 & 256 & relu6 & none & 0.5 & 0.1 & 0.001 & 0.05 & 0 & 200000 \\
Crowdflower progressive opinion & English-big & True & False & True & True & relu6 & False & 3 & 0.40 & 199 & 0.28 & 0 & 128 & relu & batch norm & 0.3 & 0.1 & 0.01 & 0.005 & 0.001 & 200000 \\
Crowdflower progressive stance & Universal-sentence-encoder & True & False & True & False & relu & True & 3 & 0.01 & 195 & 0.00 & 2 & 256 & tanh & layer norm & 0.4 & 0.005 & 0 & 0.005 & 0.0001 & 500000 \\
Crowdflower us economic performance & English-big & True & False & True & True & tanh & True & 2 & 0.31 & 53 & 0.24 & 1 & 256 & leaky relu & batch norm & 0.3 & 0.05 & 0.0001 & 0.001 & 0.0001 & 100000 \\
Customer complaint database & English-big & True & False & False & False & tanh & False & 2 & 0.03 & 69 & 0.10 & 1 & 256 & leaky relu & layer norm & 0.1 & 0.05 & 0.0001 & 0.05 & 0.001 & 1000000 \\
News aggregator dataset & Universal-sentence-encoder & False & True & True & False & sigmoid & True & 2 & 0.00 & 156 & 0.29 & 3 & 256 & relu & batch norm & 0.05 & 0.05 & 0 & 0.5 & 0.0001 & 1000000 \\
Sms spam collection & English-wiki-small & True & True & True & True & leaky relu & False & 3 & 0.20 & 54 & 0.00 & 1 & 128 & leaky relu & batch norm & 0 & 0.1 & 0 & 0.05 & 0.01 & 1000000 \\
\bottomrule
\end{tabular}
\end{adjustbox}
\caption{Search space parameters (see Table~\ref{tab:hub_ss}) for the AutoML baseline models that were selected.}
\label{tab:hub_model_params}
\end{table*}

\section{Learning Rate Robustness}

\begin{figure}[t]
\centering
\includegraphics[width=0.5\linewidth]{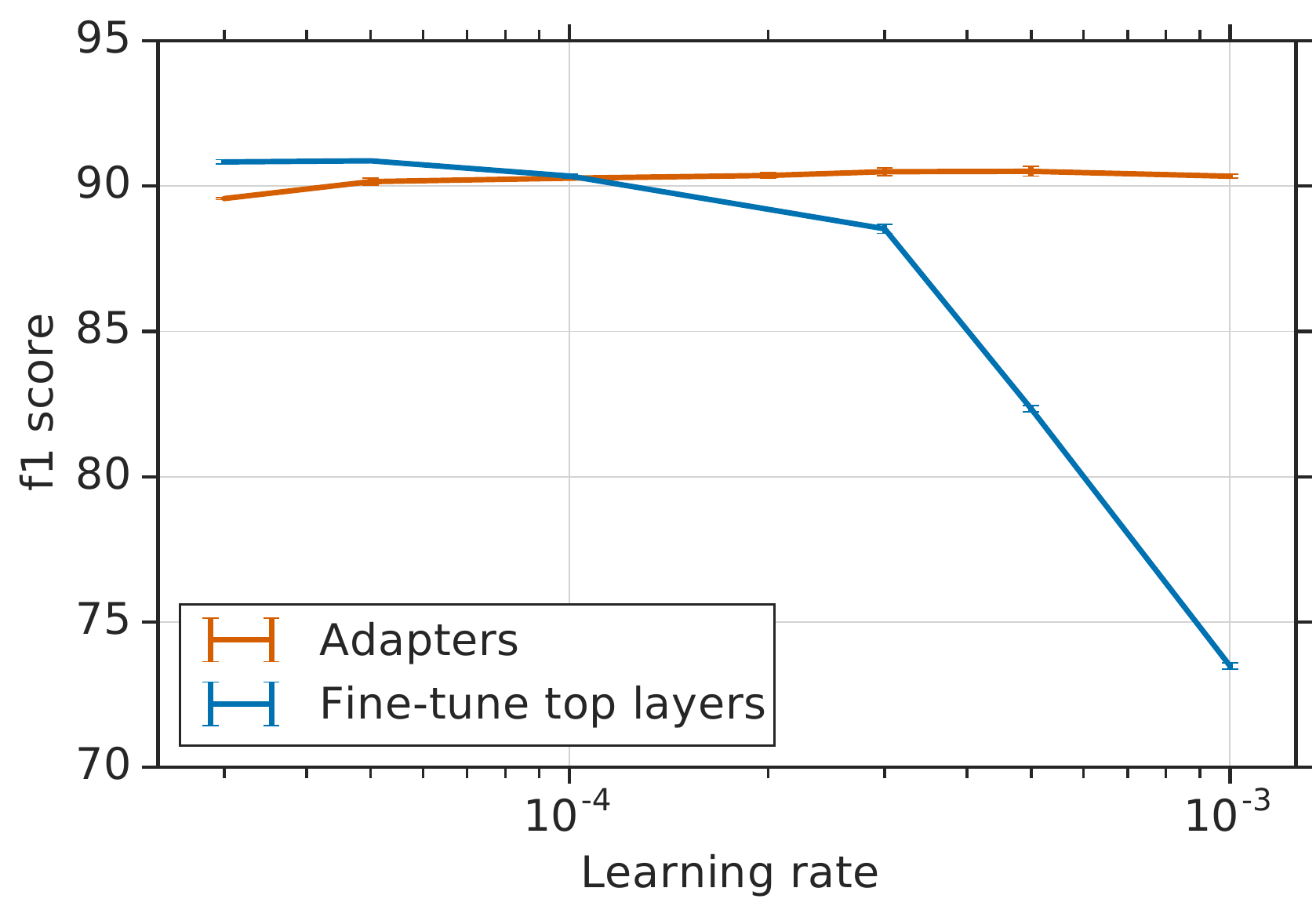}
\caption{
Best performing models at different learning rates.
Error vars indicate the s.e.m. across three random seeds.
}
\label{fig:lr}
\end{figure}

We test the robustness of adapters and fine-tuning to the learning rate.
We ran experiments with learning rates in the range $[2\cdot 10^{-5},10^{-3}]$, and selected the best hyperparameters for each method at each learning rate.
Figure~\ref{fig:lr} shows the results.

\end{document}